\documentclass[11pt]{article}
\usepackage{amsmath,amsthm,amssymb,amscd}
\usepackage{xurl}
\usepackage{hyperref}
\usepackage{geometry}
\usepackage{tikz-cd}
\usepackage{booktabs} 
\usepackage{graphicx} 
\usepackage{multirow}
\usepackage{tikz}
\usepackage{booktabs}
\usetikzlibrary{calc,positioning,3d}
\usepackage[normalem]{ulem} 
\usepackage{imakeidx}
\makeindex
\usepackage[numbers]{natbib}
\usepackage{float}

\usepackage{subfigure}
\usepackage{caption}

\usepackage{makecell} 
\usepackage{todonotes}

\newtheorem{theorem}{Theorem}[section]
\newtheorem{lemma}[theorem]{Lemma}
\newtheorem{proposition}[theorem]{Proposition}
\newtheorem{corollary}[theorem]{Corollary}
\theoremstyle{definition}
\newtheorem{definition}[theorem]{Definition}
\newtheorem{example}[theorem]{Example}
\newtheorem{remark}[theorem]{Remark}

\numberwithin{equation}{section}

\title{Combinatorial Network-Based Manifold Topological Deep Learning for Image Analysis}
\author{
	Alice Wachira$^{~1}$, Xiang Liu$^{~1}$, Zhe Su$^2$, Yiying Tong$^3$, Ge Wang$^{4}$, and 
    Guo-Wei Wei\footnote{Corresponding author: Guo-Wei Wei (guowei.wei@uga.edu).~}$^{~1,5}$  \\
    $^1$Department of Mathematics, \\ University of Georgia, Athens, GA  30602, USA.\\
   $^2$Department of Mathematics and Statistics, \\ Auburn University, AL, 36849, USA.\\
    $^3$Department of Computer Science and Engineering,\\
		Michigan State University, MI 48824, USA.\\
    $^4$ Biomedical Imaging Center,\\
        Rensselaer Polytechnic Institute, NY 12180, USA. \\
		$^5$Department of Biochemistry and Molecular Biology, \\ University of Georgia, Athens, GA  30602, USA.\\
	}
	
	\date{}
\begin{document}
 \maketitle 

\begin{abstract}	    
Medical image analysis remains fundamentally challenging because of the intricate geometric and topological structures present in medical data. Conventional convolutional neural networks model images as regular Euclidean grids, limiting their ability to preserve geometric relationships and higher-order structural information. Recently, manifold topological deep learning (MTDL) has emerged as a promising paradigm that integrates deep learning with geometric and topological representations. Nevertheless, existing methods have not yet fully exploited discrete manifold structures within combinatorial complex neural networks. To bridge this gap, we introduce CNMTDL, a MTDL framework that integrates Hodge decomposition with a combinatorial attention mechanism. In our approach, medical images are represented as discrete manifolds and decomposed into three Hodge components. Features extracted from these components are concatenated and embedded into a combinatorial complex architecture, enabling enhanced higher-order message passing between $0$-cells and $2$-cells through attention-based blocks. We evaluate CNMTDL on six two-dimensional and three-dimensional datasets from the MedMNIST v2 benchmark, demonstrating its effectiveness for medical image analysis.
\end{abstract}



\clearpage
In recent years, deep learning has emerged as a popular approach in medical image analysis \cite{zhou2021review}. The development of this approach has been driven by advances in neural network architectures, the availability of large scale datasets, and the availability of high performance computers \cite{kebaili2023deep}. In medical image tasks, deep learning models have shown great capability in detecting and classifying pathological patterns in various modalities \cite{li2023medical}. A central component of these tasks is medical image classification, which focuses on differentiating images based on modality specific or pathology-driven criteria \cite{yang2025diffmic}. Despite these achievements, medical image analysis continues to be a challenge. The complexity of biomedical science, which includes multiple imaging modalities, dataset scales, and diverse task requirements, hinders the efficiency of conventional deep learning approaches \cite{yang2023medmnist}. High-dimensional imaging data from different modalities such as Computed Tomography (CT), Magnetic Resonance Angiography (MRA), and Magnetic Resonance Imaging (MRI), further magnify these challenges due to variability in scale \cite{singh2023topological}. 

Recent efforts have been directed towards integrating topological structures into deep learning models to overcome these challenges. Since its introduction in 2017 \cite{cang2017topologynet}, Topological Deep Learning (TDL) has gained popularity due to its ability to incorporate geometric and topological structures into modern deep learning models \cite{zia2024topological}. TDL has emerged as a new paradigm in the fields of machine learning and data science \cite{papamarkou2024position}. TDL utilizes data analysis of complex high-dimensional data and integrates topological features of the data into deep learning models \cite{chen2023topological}. 

Image data can be modeled as differentiable manifolds, making it possible to design geometry- and topology-preserving algorithms. Differential topology has been proven to provide a natural mathematical technique to capture geometric and topological invariants in image data \cite{su2025topological}. In  years,  evolutionary de Rham-Hodge theory \cite{chen2021evolutionary}, and persistent de Rham-Hodge Laplacians \cite{su2024persistent} have been developed to offer mathematically rigorous frameworks for analyzing the geometric and topological structures of data on manifolds using differential forms. Combinatorial objects such as cubical complexes are also used with persistent homology to model image data \cite{wang2016object} and are capable of extracting topological characteristics from manifold data \cite{su2025topological}. However, combinatorial objects are insufficient to characterize the rich geometric and differential structure present in medical image data \cite{liu2026manifold}.

To address this challenge, researchers have recently proposed combinatorial complexes as a broad higher-order topological domain that incorporates simplicial complexes, cell complexes \cite{hajij2020cell} and graphs within a single framework \cite{hajij2022topological}. Combinatorial Complex Neural Networks (CCNN) have been designed as neural architectures defined directly on combinatorial complexes \cite{papillon2025topotune}. However, the integration of discrete manifold-based Hodge theory with combinatorial complex attention for image analysis remains unexplored.

In this work, we develop a topology-preserving deep learning setup based on discrete manifold Hodge approach combined with a combinatorial graph attention mechanism for biomedical image analysis. We introduce CNMTDL (Combinatorial Network-Based Manifold Topological Deep Learning), a model designed to bridge discrete differential geometry and topological deep learning. The proposed model combines Hodge theory, combinatorial cell neighborhood aggregation, convolution patch embedding and graph attention feed-forward blocks. Compared with earlier manifold topological deep learning approach \cite{liu2026manifold},  this provides a novel pipeline of grid embedded TDL for medical images. CNMTDL  performs image classification tasks in three phases: decomposition, combinatorial, and classification. During the decomposition phase, CNMTDL processes a biomedical image, interpreting it as a manifold defined on a Cartesian grid. A discrete Hodge decomposition is then applied to separate the vector field into curl-free, divergence-free and harmonic components, resulting in topology-aware features. For the combinatorial phase, the decomposed features are mapped onto cells induced by the image grid. Patch embeddings are used to generate feature tokens, which are organized as nodes (0-cells) on a combinatorial graph. Graph attention and feedforward blocks are used to perform message passing across neighboring cells. In addition, a 2-cell branch is implemented to capture square level interactions and both branches fused together to aggregate the multi-cell information. Finally, the resulting features are pooled and passed to a classification layer for predictions. We perform extensive experiments on a subset of the MedMNIST v2 dataset to validate the effectiveness of the proposed CNMTDL model. 


\section{Results} \label{sec:results}

\subsection{Overview of CNMTDL}\label{subsec:overview}
\begin{figure}[H]
    \includegraphics[width=1.0\linewidth]{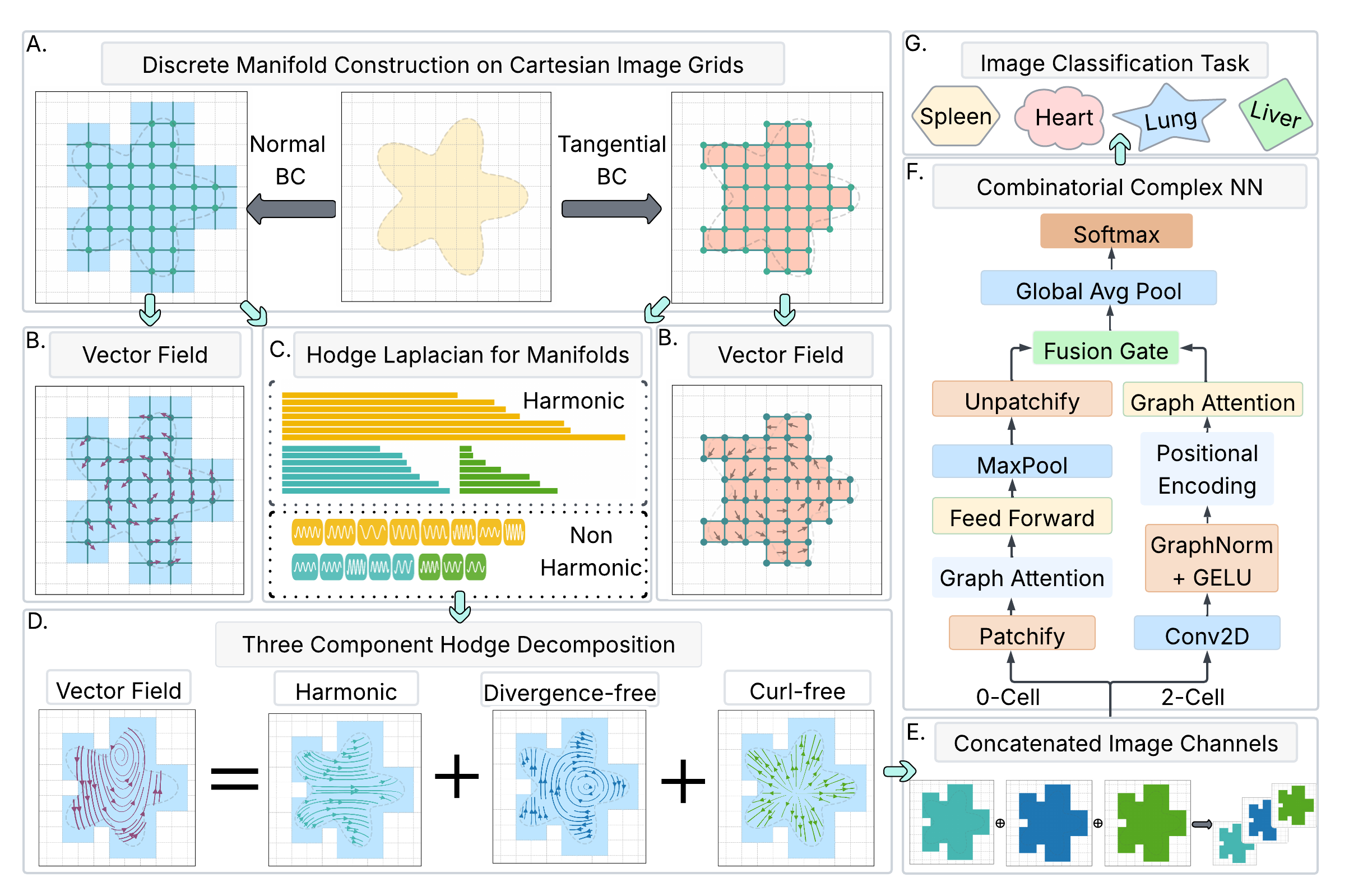}
    \caption{Workflow of the CNMTDL architecture. (A) The input image is modeled as a discrete manifold on a Cartesian grid with prescribed normal and tangential boundary conditions. (B) A vector field is constructed on the manifold representation. (C) The discrete Hodge Laplacian separates the vector field into harmonic and non-harmonic spectral components. (D) Hodge decomposition yields curl-free, divergence-free and harmonic components which are concatenated to form a new image representation (E). (F) The resulting multi-channel representation is passed to a CCNN to perform the image classification (G).} 
    \label{fig:flowchart}
\end{figure}

CNMTDLL combines the discrete manifold Hodge decomposition with the Combinatorial Complex Neural Network (CCNN) to provide a topology-preserving framework for medical image analysis. The proposed architecture consists of three major phases: decomposition (phase 1), combinatorial (phase 2) and analysis (phase 3). In phase 1, each medical image is interpreted as a discrete differential manifold on a Cartesian grid, as illustrated in Fig.~\ref{fig:flowchart}A. The intensities of each image are transformed into discrete vector fields using the finite difference method. The resultant vector field is defined over each discrete manifold with normal and tangential boundary conditions. The two boundary conditions criteria are described in Sec.~\ref{subsec: hodge decomp on images}. In Fig.~\ref{fig:flowchart}C, the Hodge Laplacian is utilized to identify the harmonic spectra of the discrete manifold. The Hodge decomposition is then applied to partition corresponding vector fields into three orthogonal components: curl-free, divergence-free and harmonic as shown in Fig.~\ref{fig:flowchart}D. Geometric and topological information of each image is revealed through this decomposition. The three decomposed components are concatenated to form a new image representation with multiple channels.

The various channels are fed into a CCNN architecture as depicted in Fig.~\ref{fig:flowchart}F in the second phase. CCNN performs message passing on cell neighborhoods and consists of two branches: 0-cell and 2-cell. A convolutional embedding layer is used in the 0-cell to partition the decomposed input image into non-overlapping patches. A neighborhood adjacency graph is created, and an attention layer is applied to relay messages among neighbors in each patch. This allows the model to capture local interactions in each section of the image. The feed-forward process then follows together with normalization. The updated tokens are reshaped and patched back into a spatial feature map. The 2-cell models square level interactions by including higher order structural relationships that are not captured at the patch level. On the induced 2-cell lattice, a four neighbor adjacency graph is created, and graph attention is used to propagate information across neighboring square cells. A fusion gate aggregates information from both cells, allowing multi-cell feature collection as illustrated in Fig.~\ref{fig:flowchart}F. The final phase involves passing the generated features into a fully connected classifier to perform the medical image classification task.

\subsection{Datasets}\label{subsec:dataset}

To assess the performance of our model, we use a subset of MedMNIST v2 dataset \cite{yang2023medmnist}. The benchmark comprises of standardized biomedical images drawn from various medical imaging domains such as X-ray, ultrasound, magnetic resonance angiography, electron microscopy, optical coherence tomography and computed tomography. The complete MedMNIST v2 benchmark has twelve 2D and six 3D datasets. In this work, we evaluate CNMTDL on six selected 2D datasets and six 3D datasets. The MedMNIST v2 dataset covers a broad spectrum of tasks such as binary and multi-class classification, ordinary prediction, and multi-label classification \cite{yang2023medmnist}. The dataset sizes vary from small to large settings of $100$ to $100,000$ \cite{yang2023medmnist}. The six 2-dimensional subset datasets include RetinaMNIST, DermaMNIST, BloodMNIST, OrganAMNIST, OrganCMNIST and OrganSMNIST while the 3-dimensional datasets include FractureMNIST3D, NoduleMNIST3D, OrganMNIST3D, SynapseMNIST3D and VesselMNIST3D. To ensure fair comparison, MedMNIST v2 provides a predefined $7:2:1$ training, validation and test splits for the datasets. For training and evaluation, the input resolutions are set to $224 \times 224$ for 2D images and $64 \times 64 \times 64$ for 3D images.

\subsection{Performance Analysis}\label{subsec:performance analysis}
We evaluate our proposed model against state of the art models using Medmnist v2 datasets based on AUC and ACC metrics as summarized in Table~\ref{tab:benchmark_results_selected} and Table~\ref{tab:benchmark_results_selected_3D}. CNMTDL consistently shows competitive performance on the evaluated datasets when compared to existing models. In particular, CNMTDL achieves the highest overall accuracy on RetinaMNIST ($0.685$), BloodMNIST ($0.991$), OrganCMNIST ($0.936$), OrganAMNIST ($0.959$), FractureMNIST3D ($0.579$), AdrenalMNIST3D ($0.879$), OrganMNIST3D ($0.930$) and SynapseMNIST3D ($0.855$). 

\begin{table}[H]
\centering
\small
\caption{Comparison of our proposed model with previous benchmark models on selected MedMNIST v2 2D datasets using ACC and AUC. Baseline results are derived from previous studies \cite{yang2023medmnist,manzari2023medvit,bai2025label,singh2025improving,rahman2025nqnn,wang2025medkaformer,zhemchuzhnikov2024ilpo,liu2022feature}. Best results are presented in bold.}
\label{tab:benchmark_results_selected}
\resizebox{\textwidth}{!}{
\begin{tabular}{lcccccccccccc}
\toprule
\multirow{2}{*}{Models}
& \multicolumn{2}{c}{Retina}
& \multicolumn{2}{c}{Derma}
& \multicolumn{2}{c}{Blood}
& \multicolumn{2}{c}{OrganC}
& \multicolumn{2}{c}{OrganS}
& \multicolumn{2}{c}{OrganA} \\
\cmidrule(lr){2-3}
\cmidrule(lr){4-5}
\cmidrule(lr){6-7}
\cmidrule(lr){8-9}
\cmidrule(lr){10-11}
\cmidrule(lr){12-13}
& AUC & ACC & AUC & ACC & AUC & ACC & AUC & ACC & AUC & ACC & AUC & ACC \\
\midrule

ResNet-18 (28) \cite{he2016deep}
& 0.717 & 0.524 
& 0.917 & 0.735
& 0.998 & 0.958 
& 0.992 & 0.900 
& 0.972 & 0.782 
& 0.997 & 0.935 \\

ResNet-50 (28) \cite{he2016deep} 
& 0.726 & 0.528 
& 0.913 & 0.735
& 0.997 & 0.956 
& 0.992 & 0.905 
& 0.972 & 0.770 
& 0.997 & 0.935 \\

ResNet-18 (224) \cite{he2016deep}
& 0.710 & 0.493 
& 0.920 & 0.754
& 0.998 & 0.963 
& 0.994 & 0.920 
& 0.974 & 0.778 
& \textbf{0.998} & 0.951 \\

ResNet-50 (224) \cite{he2016deep}
& 0.716 & 0.511 
& 0.912 & 0.731
& 0.997 & 0.950 
& 0.993 & 0.911 
& 0.975 & 0.785 
& \textbf{0.998} & 0.947 \\

auto-sklearn \cite{feurer2015efficient}  
& 0.690 & 0.515 
& 0.902 & 0.719
& 0.984 & 0.878 
& 0.976 & 0.829 
& 0.945 & 0.672 
& 0.963 & 0.762 \\

AutoKeras \cite{jin2019auto}      
& 0.719 & 0.503 
& 0.915 & 0.749
& 0.998 & 0.961 
& 0.990 & 0.879 
& 0.974 & \textbf{0.813} 
& 0.994 & 0.905 \\

Google AutoML \cite{bisong2019building}   
& 0.750 & 0.531 
& 0.914 & 0.768
& 0.998 & 0.966 
& 0.988 & 0.877 
& 0.964 & 0.749 
& 0.990 & 0.886 \\

MedVIT-L (224) \cite{manzari2023medvit} 
& 0.754 & 0.552 
& 0.920 & 0.773
& 0.996 & 0.954 
& 0.994 & 0.922 
& 0.973 & 0.806 
& 0.997 & 0.943 \\

MedVIT-T (224) \cite{manzari2023medvit}
& 0.752 & 0.534 
& 0.914 & 0.768
& 0.996 & 0.950 
& 0.991 & 0.901 
& 0.972 & 0.789 
& 0.995 & 0.931 \\

MedVIT-S (224) \cite{manzari2023medvit} 
& 0.773 & 0.561 
& 0.937 & 0.780
& 0.997 & 0.951 
& 0.993 & 0.916 
& \textbf{0.987} & 0.805 
& 0.996 & 0.928 \\

ViT \cite{dosovitskiy2020image}           
& 0.750 & 0.565 
& 0.914 & 0.745
& -- & -- 
& 0.976 & 0.835 
& 0.939 & 0.657 
& 0.978 & 0.830 \\

FPViT \cite{liu2022feature}           
& 0.753 & 0.568 
& 0.923 & 0.766
& -- & -- 
& 0.993 & 0.903 
& 0.976 & 0.785 
& 0.997 & 0.935 \\

ResNet \cite{he2016deep} + ViT \cite{dosovitskiy2020image}      
& 0.740 & 0.548 
& 0.906 & 0.748
& -- & -- 
& 0.991 & 0.900 
& 0.971 & 0.783 
& 0.995 & 0.929 \\

unORANIC \cite{doerrich2023unoranic}       
& 0.691 & 0.530 
& 0.776 & 0.699
& 0.977 & 0.848 
& -- & -- 
& -- & -- 
& -- & -- \\

BSDA \cite{zhu2024bsda}            
& 0.750 & 0.533 
& 0.931 & 0.764
& \textbf{0.999} & 0.988 
& -- & -- 
& -- & -- 
& -- & -- \\

LTP \cite{bai2025label}
& -- & 0.575
& -- & 0.787
& -- & 0.967
& -- & 0.906
& -- & 0.796
& -- & -- \\

NQNN \cite{rahman2025nqnn} 
& -- & -- 
& -- & \textbf{0.827}
& -- & 0.737 
& -- & 0.799
& -- & --
& -- & 0.814\\

SGDGCN \cite{singh2025improving}
& 0.782 & 0.550
& 0.919 & 0.771
& 0.994 & 0.940 
&0.989 & 0.880
& 0.969 & 0.784
& 0.977 & 0.850\\

MedKAFormer \cite{wang2025medkaformer}
& 0.729 & 0.535 
& 0.919 & 0.748
& 0.998 & 0.965
& 0.991 & 0.899
& 0.986 & 0.788
& 0.995 & 0.922\\

\textbf{CNMTDL}  
& \textbf{0.878} & \textbf{0.685} 
&    \textbf{0.955}  & 0.809
& \textbf{0.999} & \textbf{0.991} 
& \textbf{0.997} & \textbf{0.936} 
& 0.986 & 0.812 
& \textbf{0.998} & \textbf{0.959} \\

\bottomrule
\end{tabular}

}
\end{table}

\begin{table}[htbp]
\centering
\small
\caption{Comparison of our proposed model with previous benchmark models on selected MedMNIST v2 3D datasets using ACC and AUC. Baseline results are derived from previous studies \cite{yang2023medmnist, zhu2024bsda, liu2022feature, zhemchuzhnikov2024ilpo, zheng2023complex}. Best results are presented in bold.}
\label{tab:benchmark_results_selected_3D}
\resizebox{\textwidth}{!}{
 \begin{tabular}{lcccccccccccc}
\toprule
\multirow{2}{*}{Models}
& \multicolumn{2}{c}{Fracture}
& \multicolumn{2}{c}{Adrenal}
& \multicolumn{2}{c}{Vessel}
& \multicolumn{2}{c}{Organ}
& \multicolumn{2}{c}{Nodule}
& \multicolumn{2}{c}{Synapse} \\
\cmidrule(lr){2-3}
\cmidrule(lr){4-5}
\cmidrule(lr){6-7}
\cmidrule(lr){8-9}
\cmidrule(lr){10-11}
\cmidrule(lr){12-13}
& AUC & ACC & AUC & ACC & AUC & ACC & AUC & ACC & AUC & ACC & AUC & ACC \\
\midrule

ResNet-18 \cite{he2016deep} + 2.5D \cite{yang2021reinventing}  
& 0.587 & 0.451 & 0.718 & 0.772 & 0.748 & 0.846 & 0.977 & 0.788 & 0.838 & 0.835 & 0.634 & 0.696 \\

ResNet-18 \cite{he2016deep} + 3D  \cite{yang2021reinventing}   
& 0.712 & 0.508 & 0.827 & 0.721 & 0.874 & 0.877 & \textbf{0.996} & 0.907 & 0.863 & 0.844 & 0.820 & 0.745 \\

ResNet-18 \cite{he2016deep} + ACS \cite{yang2021reinventing}  
& 0.714 & 0.497 & 0.839 & 0.754 & \textbf{0.930} & \textbf{0.928} & 0.994 & 0.900 & 0.873 & 0.847 & 0.705 & 0.722 \\

ResNet-50 \cite{he2016deep} + 2.5D \cite{yang2021reinventing} 
& 0.552 & 0.397 & 0.732 & 0.763 & 0.751 & 0.877 & 0.974 & 0.769 & 0.835 & 0.848 & 0.669 & 0.735 \\

ResNet-50 \cite{he2016deep} + 3D \cite{yang2021reinventing}  
& 0.725 & 0.494 & 0.828 & 0.745 & 0.907 & 0.918 & 0.994 & 0.883 & 0.875 & 0.847 & 0.851 & 0.795 \\

ResNet-50 \cite{he2016deep} + ACS \cite{yang2021reinventing}   
& 0.750 & 0.517 & 0.828 & 0.758 & 0.912 & 0.858 & 0.994 & 0.889 & 0.886 & 0.841 & 0.719 & 0.709 \\

auto-sklearn \cite{feurer2015efficient}   
& 0.628 & 0.453 & 0.828 & 0.802 & 0.910 & 0.915 & 0.977 & 0.814 & \textbf{0.914} & \textbf{0.874} & 0.631 & 0.730 \\

AutoKeras \cite{jin2019auto} 
& 0.642 & 0.458 & 0.804 & 0.705 & 0.773 & 0.894 & 0.979 & 0.804 & 0.844 & 0.834 & 0.538 & 0.724 \\

FPViT (224) \cite{liu2022feature}       
& 0.640 & 0.438 & 0.801 & 0.704 & 0.770 & 0.888 & 0.923 & 0.800 & 0.814 & 0.822 & 0.530 & 0.712 \\

BSDA \cite{zhu2024bsda} 
& 0.731 & 0.569 & 0.892 & 0.838 & 0.917 & 0.932 & 0.994 & 0.887 & 0.892 & 0.861 & - & - \\

ILPO-NET (average) \cite{zhemchuzhnikov2024ilpo} 
& \textbf{0.776} & 0.577 & 0.880 & 0.811 & 0.888 & 0.888 & 0.972 & 0.728 & 0.900 & 0.861 & 0.854 & 0.782 \\


\textbf{CNMTDL}     
& 0.760 & \textbf{0.579} & \textbf{0.923} & \textbf{0.879} & 0.929 & 0.901 & \textbf{0.996} &\textbf{ 0.930 } & \textbf{0.914} & 0.871 & \textbf{0.932} & \textbf{0.855} \\

\bottomrule
\end{tabular}
}
\end{table}

\begin{figure}[H]
    \centering
    \includegraphics[width=1.0\linewidth]{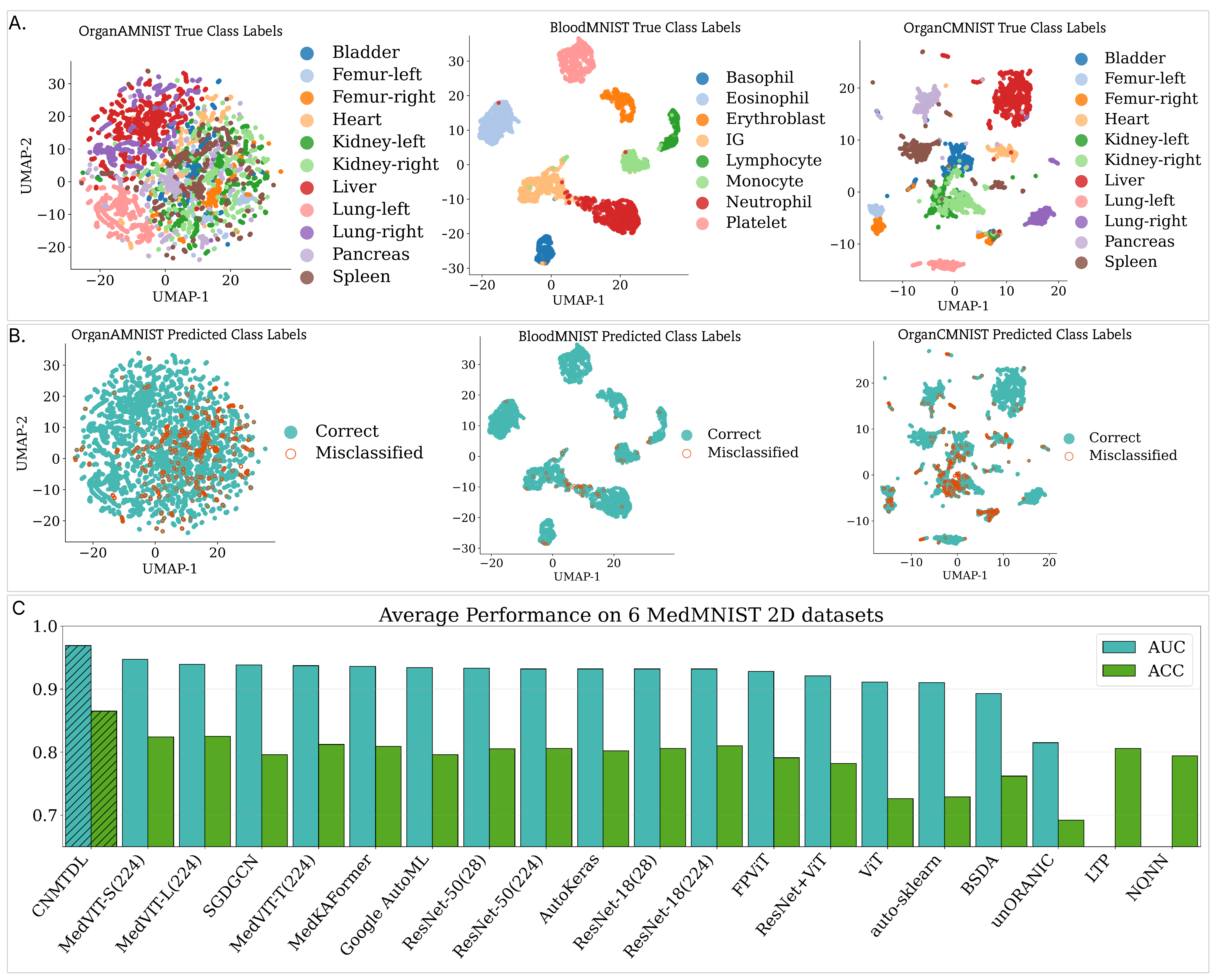}
    \caption{CNMTDL performance and embedding separability on MedMNIST 2D datasets. UMAP visualization of ground truth (A) and predicted class labels (B). (C) Average AUC and ACC across six MedMNIST v2 2D datasets for CNMTDL and benchmark models.}
    \label{fig:2D analysis and umap}
\end{figure}

Fig.~\ref{fig:2D analysis and umap}C illustrates the average ACC and AUC performance in six 2D datasets. In contrast to second best model MedVIT-S, which obtains ACC value of $0.824$ and AUC value of $0.947$, CNMTDL demonstrates the highest overall ACC value $0.865$ and AUC value $0.969$. 

\begin{figure}[H]
    \centering
    \includegraphics[width=1.0\linewidth]{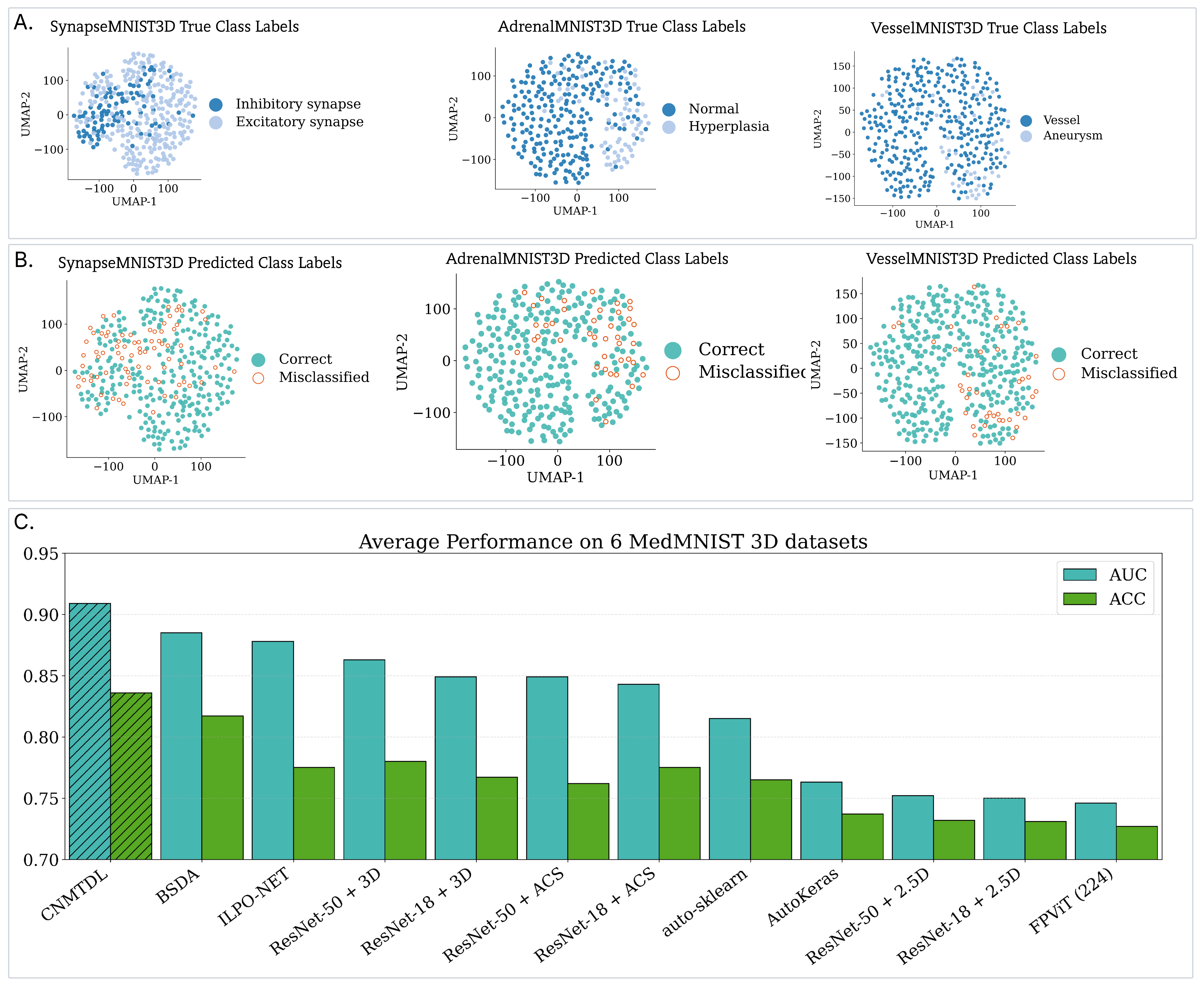}
    \caption{CNMTDL performance on MedMNIST v2 3D datasets. UMAP embeddings colored by ground truth (A) and predicted class labels (B). (C) Average AUC and ACC across six MedMNIST 3D datasets for CNMTDL and benchmark models.}
    \label{fig:average auc and acc of 6 datasets 3D}
\end{figure}

Fig.~\ref{fig:average auc and acc of 6 datasets 3D}A presents the average ACC and AUC performance across six MedMNIST v2 3D datasets. CNMTDL achieves the best overall performance with an average AUC of $0.909$ and an average ACC of $0.836$. Compared with the second best model BSDA which obtains an average AUC of $0.886$ and an average ACC of $0.818$, CNMTDL provides clear improvements in both evaluation metrics.

To assess how CNMTDL separates different classes across the various datasets, we visualize the learned embeddings using Uniform Manifold Approximation and Projection (UMAP). For selected 2D datasets we present a pair of UMAP plots with true class labels in Fig.~\ref{fig:2D analysis and umap}A and CNMTDL predicted class labels in Fig.~\ref{fig:2D analysis and umap}B. The ground truth embeddings reveal varying class separability with datasets such as BloodMNIST and OrganCMNIST displaying distinct clusters. In contrast, more complex datasets such as OrganAMNIST show significant class overlap. BloodCMNIST demonstrates the best prediction in class separation, with only minimal boundary errors, consistent with its strong classification accuracy. In datasets with inter-class mixing such as OrganAMNIST, misclassified labels tend to appear near class boundaries. For the selected 3D datasets the UMAP embeddings in Fig.~\ref{fig:average auc and acc of 6 datasets 3D} confirms that CNMTDL learns structured embedding spaces that reflect the underlying class geometry although the class separation varies across datasets. In the true label embeddings shown in Fig.~\ref{fig:average auc and acc of 6 datasets 3D}A SynapseMNIST3D, AdrenalMNIST3D and VesselMNIST3D have dense class overlap indicating that the 3D classification tasks contain very challenging regions where samples from different classes share similar structural patterns. The embeddings shown in Fig.~\ref{fig:2D analysis and umap}B and Fig.~\ref{fig:average auc and acc of 6 datasets 3D}B suggest that misclassification errors are not randomly distributed instead, they are concentrated in regions where class manifolds overlap or transition. 

\subsection{Ablation Study}
We first conduct an ablation experiment to investigate the effect of different choices of $n$-cells within the CCNN network. We train two reduced CCNN variants:  $0$-cell CCNN which uses node representations and $2$-cell CCNN incorporates square region representations. The ablation results presented in Table~\ref{tab:ablation_cell_results} confirm the essential role of each $n$-cell and highlight the importance of combining the $n$-cells in the full CNMTDL model. CNMTDL improves the accuracy across all the datasets. 

\begin{table}[H]
\centering
\caption{Comparison of CNMTDL with CCNN variants constructed using only $0$-cell or only $2$-cell frameworks.}
\label{tab:ablation_cell_results}
\resizebox{\linewidth}{!}{
\begin{tabular}{lcccccccccccccc}
\toprule
\multirow{2}{*}{Models}
& \multicolumn{2}{c}{Retina}
& \multicolumn{2}{c}{Blood}
& \multicolumn{2}{c}{OrganC}
& \multicolumn{2}{c}{OrganS} 
& \multicolumn{2}{c}{Vessel}
& \multicolumn{2}{c}{Adrenal}
& \multicolumn{2}{c}{Synapse}\\
\cmidrule(lr){2-3}
\cmidrule(lr){4-5}
\cmidrule(lr){6-7}
\cmidrule(lr){8-9}
\cmidrule(lr){10-11}
\cmidrule(lr){12-13}
\cmidrule(lr){14-15}
& AUC & ACC & AUC & ACC & AUC & ACC & AUC & ACC & AUC & ACC & AUC & ACC & AUC & ACC \\
\midrule
$0$-cell CCNN
& \textbf{0.878} & 0.640
& \textbf{0.999} & 0.989
& 0.996 & 0.932
& 0.978 & 0.797 
& 0.915  & \textbf{0.901 }
& 0.920 & 0.852 
& 0.892 & 0.838 \\

$2$-cell CCNN
& 0.854 & 0.632
& 0.998 & 0.975
& 0.994 & 0.911
& 0.973 & 0.764 
 & 0.829  & 0.893
& 0.913 & 0.873
 &  0.769 &  0.776 \\

CNMTDL
& \textbf{0.878} & \textbf{0.685}
& \textbf{0.999} & \textbf{0.991}
& \textbf{0.997} & \textbf{0.936}
&\textbf{ 0.985} & \textbf{0.812}
& \textbf{0.929} & \textbf{0.901}
& \textbf{0.923} & \textbf{0.879} 
& \textbf{0.932} & \textbf{0.855}\\
\bottomrule
\end{tabular}
}
\end{table}

We further perform an ablation experiment to evaluate the contribution of the manifold topology decomposition in CNMTDL. In the proposed model CNMTDL, each medical image is decomposed into three Hodge components. We examine the role of Hodge decomposition by designing several reduced models where we omit parts of the decomposition components. First, we consider a model that relies only on the curl-free component as an input for the CCNN architecture, which we denote by Curl-CCNN. Next, we analyze another model using only the divergence-free component, referred to as Div-CCNN. We then combine the curl-free and divergence-free components, leaving out the harmonic component, and we denote this model by Curl-Div-CCNN. The experimental results are presented in Table~\ref{tab:ablation_studies}. The overall accuracy for the bloodMNIST dataset does not change across all the models. When using only the div-free component (Div-CCNN) or the curl-free component (Curl-CCNN), the performance drops, suggesting that relying on a single component is insufficient to capture the complete structural information present in medical images. In particular, Div-CCNN model records the lowest performance indicating that the divergence-free component alone does not represent the underlying geometric of the image fully. A significant decrease in performance is observed when DIV-CCNN model is evaluated on the AdrenalMNIST3D dataset. Combining the curl-free and divergence-free components, improves the performance confirming that these features provide complementary information. The best performance is achieved by incorporating all three components in the Hodge decomposition. Our study shows that the complete Hodge decomposition provides a more effective representation of the underlying geometric and topological structures in medical images.
\begin{table}[H]
\centering
\small
\caption{Comparison of CNMTDL with models constructed using different Hodge decomposition components.}
\label{tab:ablation_studies}
\resizebox{\textwidth}{!}{
\begin{tabular}{lcccccccccccccc}
\toprule
\multirow{2}{*}{Models}
& \multicolumn{2}{c}{Retina}
& \multicolumn{2}{c}{Derma}
& \multicolumn{2}{c}{Blood}
& \multicolumn{2}{c}{OrganC}
& \multicolumn{2}{c}{OrganS}
& \multicolumn{2}{c}{OrganA} 
& \multicolumn{2}{c}{Adrenal}\\
\cmidrule(lr){2-3}
\cmidrule(lr){4-5}
\cmidrule(lr){6-7}
\cmidrule(lr){8-9}
\cmidrule(lr){10-11}
\cmidrule(lr){12-13}
\cmidrule(lr){14-15}
& AUC & ACC & AUC & ACC & AUC & ACC & AUC & ACC & AUC & ACC & AUC & ACC & AUC & ACC\\
\midrule
Curl-CCNN 
& 0.857 & 0.670
& 0.953 & 0.803
& \textbf{0.999} & \textbf{0.991} 
& 0.996 &  0.933
& 0.981 &  0.801  
& \textbf{0.998}  & 0.951
& 0.919 & 0.856 \\
Div-CCNN
& 0.820 & 0.608 
& 0.900 &  0.734
& \textbf{0.999} &  0.974
& 0.992 &  0.897
& 0.969  &  0.748
& 0.992 & 0.901  
& 0.518 & 0.769 \\

Curl-Div-CCNN
& 0.873  & 0.653 
& 0.955 & 0.802
& \textbf{0.999} &  0.990
& 0.996  &  0.930
& 0.982  &  0.810
& 0.998  & 0.956 
& 0.914  & 0.856 \\

\textbf{CNMTDL}  
& \textbf{0.878} & \textbf{0.685} 
&    \textbf{0.955}  & \textbf{0.809}
& \textbf{0.999} & \textbf{0.991} 
& \textbf{0.997} & \textbf{0.936} 
& 0.986 & 0.812 
& \textbf{0.998} & \textbf{0.959} 
&\textbf{0.923 }& \textbf{0.879} \\

\bottomrule
\end{tabular}

}
\end{table}

\section{Discussion} \label{sec:discussion}
\begin{figure}[H]
    \centering
    \includegraphics[width=1.0\linewidth]{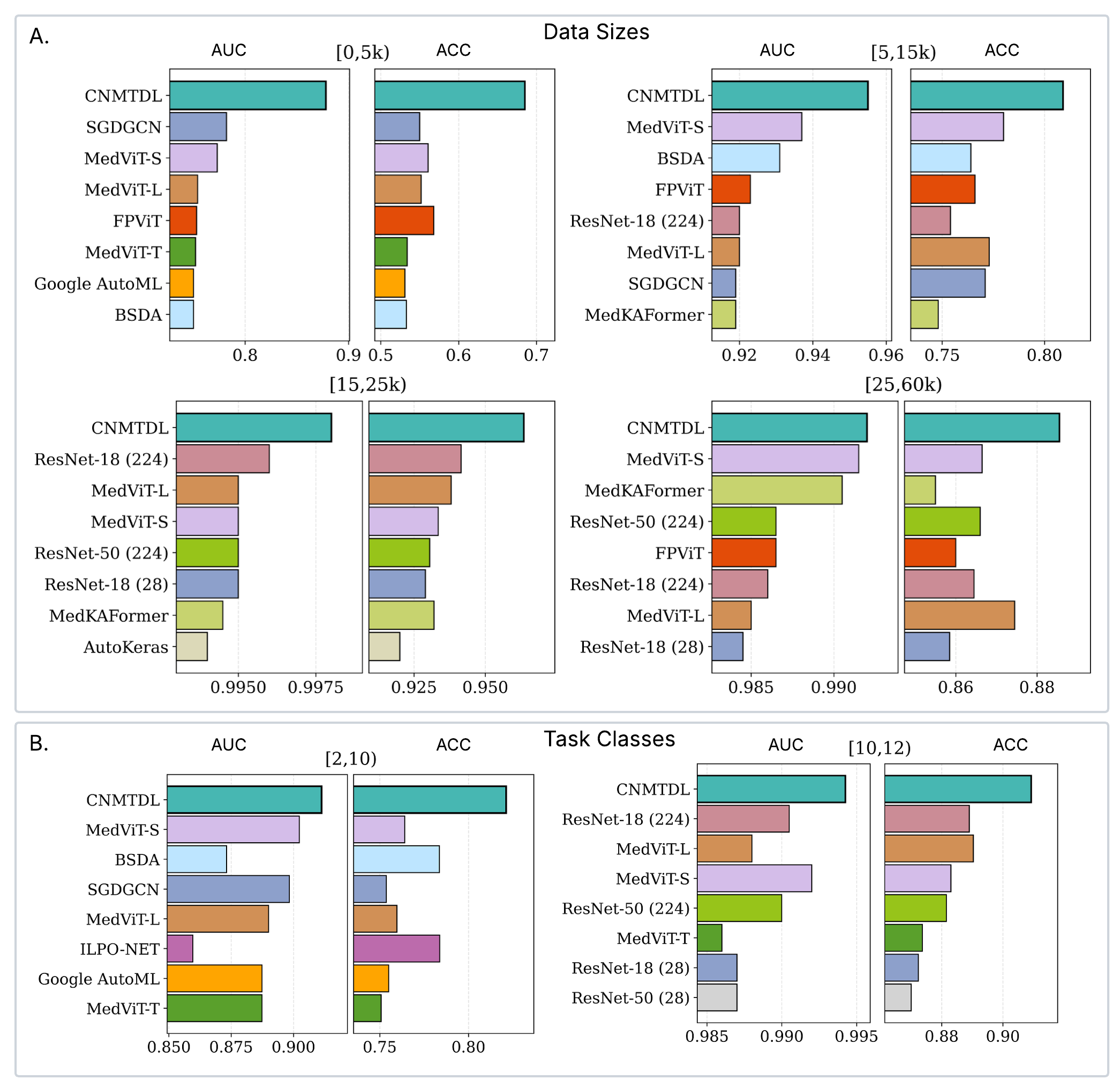}
    \caption{Performance across dataset scales and classification tasks. (A) AUC and ACC of the top performing models grouped by dataset sizes. (B) AUC and ACC of the top performing models grouped by number of classification task classes.}
    \label{fig:datascales_classes}
\end{figure}

We further evaluate the robustness of the proposed CNMTDL model by analyzing its performance across datasets with varying data scales and task complexities. In particular, the datasets are grouped according to the sample sizes and the number of classification class tasks. For data scale analysis we only consider the 2D datasets which are divided into four groups based on the number of samples: $[0,5{,}000)$, $[5{,}000,15{,}000)$, $[15{,}000,25{,}000)$, and $[25{,}000,60{,}000)$. We also examine the impact of task complexity by grouping datasets according to the number of classification classes into two ranges: $[2,10)$ and $[10,12)$. For each group, we report the top eight performing models based on AUC and ACC, as shown in Fig.~\ref{fig:datascales_classes}. The results shown in Fig.~\ref{fig:datascales_classes} highlight the robustness of CNMTDL as the model consistently maintains strong performance even under varying dataset sizes and number of task classes.

The experimental results reveal that CNMTDL performs exceptionally well on various MedMNIST v2 datasets. For the 2D datasets, the model excels in both smaller datasets like RetinaMNIST and larger ones such as OrganAMNIST. Several 3D datasets including VesselMNIST3D, NoduleMNIST3D, SynapseMNIST3D, and AdrenalMNIST3D have class imbalance, with some classes being more represented than others. This imbalance can make classification more challenging since models may become biased toward majority classes and fail to adequately learn discriminative features for minority classes. Despite these challenges, CNMTDL maintains strong performance across the 3D datasets demonstrating its robustness for volumetric medical image classification. Unlike standard convolutional architectures or pure transformer based models, CNMTDL combines Hodge decomposition with combinatorial graph attention message passing defined on multiple cell dimensions. This integrated approach allows the model to incorporate geometric and topological structures into its learning and image analysis process. 

A fundamental aspect of CNMTDL is the integration of a manifold framework and combinatorial complex learning. Medical images are first interpreted as discrete manifolds and represented as vector fields obtained from Hodge decomposition. Rather than treating images solely as arrays of intensities, this topological aspect encodes structured geometric properties before neural processing. Additionally, the model constructs a combinatorial complex from the decomposed image. The combination of $0$-cells and $2$-cells allows the network to take advantage of higher-order structural features. The combination of these aspects enables the proposed model to achieve optimal performance. 

It is worth noting that although CNMTDL achieves the best overall average performance, some competing methods obtain better results on specific datasets. In addition, C-Mixer \cite{zheng2023complex} has reported strong performance on several MedMNIST v2 3D datasets. However, we did not include C-Mixer in our main comparison due to differences in the training samples used for some 3D datasets, which makes a direct comparison less consistent with the experimental setting adopted in this work.

This work demonstrates the potential of CNMTDL in learning discriminative and structure-aware representations in both 2D and 3D medical image datasets. There are several directions that could further extend CNMTDL architecture in the future. The combinatorial complex aspect could be enhanced by incorporating multiple cell representations to capture richer interactions in decomposed images and introduce explicit cross-dimensional message passing to model interactions across cell dimensions.

\section{Methods}\label{sec:materials}

Medical images are defined on Cartesian grids and can be interpreted as discrete manifolds with boundary, making them suitable for analysis with differential geometry techniques. Previous work has shown that topology-preserving Hodge decomposition can be used to extract meaningful information from these images \cite{su2024hodge}.
\subsection{Continuous Hodge Decomposition on Closed Manifolds}\label{subsec: cont hodge}

Consider a smooth, orientable and compact manifold $M$ of dimension $m$ with boundary $\partial M$. The space of smooth differential $k$-forms on $M$, denoted as $\Omega^{k}(M)$, can be viewed as the set of smooth sections of the $k$-th exterior power of the cotangent bundle. 

The exterior derivative is a linear operator that maps $k$-forms to $(k\!+\!1)$-forms and will be denoted as $d$. We say that a differential $k$-form is closed if $d\omega=0$ and is exact if there exists a $(k\!-\!1)$-form $\xi$ such that $d\xi = \omega$.

Assuming $g$ is a Riemannian metric on the manifold $M$, the Hodge star operator $\star$ can be defined as a mapping from $k$-forms to $(m\!-\!k)$-forms (through $g$-orthogonal complement) and induces an $L^2$ inner product on the space $\Omega^k(M)$ (through wedge product with the Hodge star). The codifferential operator is an associated linear map $\delta: \Omega^k(M) \to \Omega^{k-1}(M)$ defined by
\begin{equation}
    \delta =(-1)^{m(k-1)+1}\star d \star,
\end{equation}
with its nilpotent property  $\delta \delta=0$ following that of the exterior derivative, $dd=0.$ We say that a differential form $\omega \in \Omega^{k}(M)$ is coclosed if $\delta \omega=0$ and coexact if there  exists a $(k\!+\!1)$-form $\xi$ such that $\omega = \delta \xi.$

As a second order differential operator acting on differential forms, the Hodge Laplacian is defined as $\Delta = d\delta + \delta d.$ The differential operator $d$ and codifferential $\delta$ satisfy the condition that 
\begin{equation}
    (d\omega,\nu)- (\omega, \delta \nu) = \int_{\partial M} \omega \wedge  \star \nu,
\end{equation}
where $\nu$ is a $(k\!+\!1)$-form, $\omega$ is a $k$-form and $\wedge$ is the wedge product. Thus, these two operators $d$ and $\delta$ are adjoint if $M$ is a closed manifold (i.e., a compact manifold without boundary). A $k$-form $\omega$ is said to be harmonic if it lies in the kernel of the Hodge Laplacian, i.e., $\Delta \omega=0.$

Let $\mathcal{H}^{k}_{\Delta}(M)$ be the space of harmonic $k$-forms, which can also be defined as the finite-dimensional kernel of the Hodge Laplacian when $M$ is a closed manifold. The Hodge decomposition theorem states that the space of smooth differential $k$-forms on $M$ can be expressed as an orthogonal decomposition \cite{zhao20193d},
\begin{equation}
    \Omega^k(M)= d\Omega^{k-1} \oplus \delta \Omega^{k+1} \oplus \mathcal{H}^{k}_{\Delta}(M).
\end{equation}

When the manifold has a boundary, the orthogonality of the Hodge Decomposition holds only when suitable boundary conditions are imposed. The commonly used normal and tangential boundary conditions have physical interpretations and define two subspaces 
\begin{equation}
    \Omega^{k}_{n}(M)=\{\omega \in \Omega^{k}(M)\,|\,\omega\big|_{\partial M}=0\}, \quad \Omega^{k}_{t}(M)=\{\omega \in \Omega^{k}(M)\,|\,\star \omega\big|_{\partial M}=0\}.
\end{equation}
The Hodge star operator induces an isomorphism between the space of normal $k$-forms $\Omega^{k}_{n}(M)$ and the space of tangential $(m\!-\!k)$-forms $\Omega^{m-k}_{t}(M)$. Moreover, the exterior derivative preserves the normal boundary conditions while the codifferential preserves the tangential boundary conditions. Under these boundary conditions, the Hodge-Morrey decomposition \cite{morrey1956variational} provides a way to decompose $\Omega^k(M)$ orthogonally into three subspaces 
\begin{equation}\label{eq: morrey decomposition}
    \Omega^{k}(M)=d\Omega^{k-1}_{n}(M) \oplus \delta \Omega_{t}^{k+1}(M) \oplus \mathcal{H}^{k}(M),
\end{equation}
where $\mathcal{H}^{k}(M)=\ker d \cap \ker \delta$ denotes the space of closed and coclosed differential $k$-forms. 

Two ways to orthogonally decompose the space $\mathcal{H}^{k}(M)$ were proposed by Friedrichs \cite{friedrichs1955differential} that is 
\begin{equation}\label{eq: harm spaces}
    \mathcal{H}^{k}(M) = \mathcal{H}_{n}^{k}(M) \oplus (d\Omega^{k-1}(M) \cap \delta \Omega^{k+1}(M)) \text{ or }  \mathcal{H}_{t}^{k}(M) \oplus (d\Omega^{k-1}(M) \cap \delta \Omega^{k+1}(M)),
\end{equation}
where $\mathcal{H}_{n}^{k}(M)=\mathcal{H}^{k}(M) \cap \Omega^{k}_{n}(M)$ and $\mathcal{H}_{t}^{k}(M)=\mathcal{H}^{k}(M) \cap \Omega^{k}_{t}(M)$ are not necessarily orthogonal with respect to the $L^2$ inner product. However, if $M$ is a compact Euclidean domain, $\mathcal{H}_{n}^{k}(M)$ and $\mathcal{H}_{t}^{k}(M)$ are orthogonal \cite{shonkwiler2009poincare}. In such cases, \ref{eq: harm spaces} yields a decomposition with three components 
\begin{equation}\label{eq: friedrichs decomposition}
    \mathcal{H}^{k}(M) = (\mathcal{H}_{n}^{k}(M) \oplus \mathcal{H}_{t}^{k}(M)) \oplus (d\Omega^{k-1}(M) \cap \delta \Omega^{k+1}(M)).
\end{equation}
From (\ref{eq: morrey decomposition}) and (\ref{eq: friedrichs decomposition}), we obtain Hodge-Morrey-Friedrichs decomposition consisting of five components
\begin{equation}\label{eq: Hodge friedrichs decomposition}
    \Omega^{k}(M)=d\Omega^{k-1}_{n}(M) \oplus \delta \Omega_{t}^{k+1}(M) \oplus (\mathcal{H}_{n}^{k}(M) \oplus \mathcal{H}_{t}^{k}(M)) \oplus (d\Omega^{k-1}(M) \cap \delta \Omega^{k+1}(M)).
\end{equation}
As a result of the Hodge-Morrey-Friedrichs decomposition(\ref{eq: Hodge friedrichs decomposition}), for every differential $k$-form $w\in \Omega^{k}(M)$, there is a unique orthogonal decomposition with five components
\begin{equation}\label{eq: 5 component decomposition}
    \omega = d \theta_{n} + \delta \mu_{t}+h_n +h_t +\nu,
\end{equation}
where $\theta_{n}\in \Omega_{n}^{k-1}(M)$, $\mu_{t}\in \Omega_{n}^{k+1}(M)$, $h_n\in \mathcal{H}_{n}^{k}(M)$, $h_t\in \mathcal{H}_{t}^{k}(M)$  and $\nu \in d\Omega^{k-1}(M) \cap \delta \Omega^{k+1}(M)$.

The first two components of the decomposition(\ref{eq: 5 component decomposition}) can be computed by determining the corresponding potentials $\theta_n \in \Omega^{k-1}_{n}$ and $\mu_{t} \in \Omega_{t}^{k+1}$, and then applying the differential $d$ and codifferential $\delta$ respectively. It should be noted that these potentials $\theta_{n}$ and $\mu_{t}$ are not uniquely determined since $d\theta_n=d(\theta_{n}+d\nu)$ and $\delta \mu_{t}=\delta(\mu_{t}+\delta \mu)$ for arbitrary $\nu \in \Omega_{n}^{k-2}(M)$ and $\mu \in \Omega_{t}^{k+2}(M)$.

For efficiency and stability of linear solvers, the uniqueness of these potentials are crucial, which can be enforced by imposing the gauge conditions $\delta \theta_n=0$ and $d\mu_{t}=0$ and fixing their projection on $\mathcal{H}_{t}(M)$ and $\mathcal{H}_{n}(M)$. With the gauge conditions, we obtain the following
\begin{subequations}\label{eq: gauge conditions}
\begin{align}
\delta \omega 
&= (\delta d + d\delta)\theta_n 
= \Delta \theta_n,
\label{eq: gauge a} \\[6pt]
d \omega 
&= (\delta d + d\delta)\mu_t 
= \Delta \mu_t.
\label{eq: gauge b}
\end{align}
\end{subequations}
The potential $\theta_n$ can now be obtained by solving (\ref{eq: gauge a}) subject to the boundary conditions $\theta_{n}\big|_{\partial M}=0$ and $\delta \theta_{n}\big|_{\partial M}=0$ and similarly $\mu_t$ obtained by solving (\ref{eq: gauge b}) subject to boundary conditions $\star \mu_{t}\big|_{\partial M}=0$ and $\star d\mu_{t}\big|_{\partial M }=0.$ 

Since the harmonic spaces $\mathcal{H}_{n}^{k}(M)$ and $\mathcal{H}_{t}^{k}(M)$ are of finite dimension, $h_n$ and $h_t$ are directly computed by projecting $\omega$ onto $\mathcal{H}_{n}^{k}(M)$ and $\mathcal{H}_{t}^{k}(M)$ using an orthogonal basis. Therefore, the last component $\nu$ becomes $\nu = \omega -d\theta_{n}-\delta \mu_{t}-h_{n}-h_{t}.$ In this study, we will restrict our attention to the three component decomposition 
\begin{equation}\label{eq: three components}
    \omega = d \theta_{n} + \delta \mu_{t}+h,
\end{equation}
where $h=h_n+h_t+\nu.$

\subsection{Topological Discrete Hodge Decomposition for Medical Images}\label{subsec: hodge decomp on images}
A discrete manifold $M$ can be represented on a two-dimensional Cartesian grid, as a sublevel set of a level set function defined on the grid. In polygonal and simplicial discretizations, the boundary elements can be identified explicitly, simplifying the construction of projection matrices onto the boundary or interior of the domain. On the other hand, applying boundary conditions through projection matrices becomes complex when a discrete manifold $M$ is defined on a sublevel set of functions within the Cartesian grid. To address this challenge, a grid based approach that modifies the discrete Hodge star can be implemented on $M$ as explained by \cite{su2024hodge2}. 

On the Cartesian grid, $M$ can be described in terms of $k$-cells. Let $I_m$ denote a Cartesian grid whose cells are oriented consistently with the coordinate axes. Grid vertices, edges, and square faces correspond to $0$-cells, $1$-cells, and $2$-cells, respectively. Each $k$-cell can be viewed as a $k$-dimensional hypercube. An image intensity function can be interpreted as a discrete $0$-form as it assigns scalar values to grid vertices. 

In our work, we adopt the approach used in \cite{ribando2024combinatorial,liu2026manifold}, which determines the boundary of $M$ under two types of boundary conditions: normal (Dirichlet) and tangential (Neumann) conditions. For normal boundary conditions, grid cells are included if they contain at least one vertex belonging to $M$, while for tangential boundary conditions, cells are included only if at least one vertex of the associated dual belongs to $M.$ The selection from the former and the latter defines two distinct collections of cells, namely normal support and tangential support, respectively. In general, these supports differ, unlike in discretizations based on meshes, and neither is guaranteed to be a superset of the other.

Restricting computations on appropriate supports, we define projection matrices $P_k$ that act on cochains. The matrices $P_{k,n}$ and $P_{k,t}$  map $k$-cochains to their restriction on normal and tangential supports respectively. These matrices are obtained from the identity matrix by deleting rows corresponding to $k$-cells that lie outside the chosen support. The exterior derivative and Hodge star on $M$ can be discretized and denoted by $D_k$ and $S_k$ respectively. The resulting discrete operators for the former and latter on $I_m$ under normal and tangential conditions are given by
\begin{subequations}
    \begin{align}
        &D_{k,n}=P_{k+1,n}D_{k}P_{k,n}^{\top} \quad D_{k,t}=P_{k+1,t}D_{k}P_{k,t}^{\top},\\
        &S_{k,n}=P_{k+1,n}S_{k}P_{k,n}^{\top} \quad S_{k,t}=P_{k+1,t}S_{k}P_{k,t}^{\top},
        \end{align}
\end{subequations}
respectively. 
Similarly, the discrete codifferential for the two boundary conditions can be expressed as $S_{k-1,n}^{-1}D_{k-1,n}^{\top}S_{k,n}$ and $S_{k-1,t}^{-1}D_{k-1,t}^{\top}S_{k,t}$. When the continuous operators $d$ and $\delta$ in $\Delta=d\delta + \delta d$ are replaced with their corresponding discrete matrices, they yield a non-symmetric matrix\cite{su2024hodge2}. A symmetric formulation is obtained by defining $L_{k}$, the discrete Hodge Laplacian as a discrete counterpart of the operator $\star \Delta$, 
\begin{equation}
    L_{k} = D_{k}^{\top}S_{k+1}D_{k} + S_{k}D_{k-1}S_{k-1}^{-1}D_{k-1}^{\top}S_{k}.
\end{equation}
The discrete Laplacians corresponding to the normal and tangential boundary conditions are given by 
\begin{subequations}
    \begin{align}
       & L_{k,n} = D_{k,n}^{\top} S_{k+1,n}D_{k,n} + S_{k,n}D_{k-1,n}S_{k-1,n}^{-1}D_{k-1,n}^{\top}S_{k,n},\\
    &L_{k,t} = D_{k,t}^{\top} S_{k+1,t}D_{k,t} + S_{k,t}D_{k-1,t}S_{k-1,t}^{-1}D_{k-1,t}^{\top}S_{k,t},
    \end{align}
\end{subequations}

The normal and tangential harmonic components have a direct topological interpretation. Every normal harmonic field is linked to an equivalence class in the relative de Rham cohomology. The tangential harmonic field is associated with the (absolute) de Rham cohomology. As a result, the dimensions of $\mathcal{H}_{n}^{k}(M)$ and $\mathcal{H}_{t}^{k}(M)$ are determined by the topology of $M$, given by
\begin{subequations}
    \begin{align}
    & \dim \mathcal{H}_{n}^{k}(M) =\beta_{m-k}=\dim (\ker L_{k,n}),\\
    & \dim \mathcal{H}_{t}^{k}(M) =\beta_{k}=\dim (\ker L_{k,t}),
\end{align}
\end{subequations}
where $\beta_{k}$ is the $k^\text{th}$ Betti number of $M$.

Given an RGB input image with three channels we construct a vector-field representation for each channel separately and decompose each representation into curl-free, divergence-free and harmonic components. Each image channel is converted into a discrete vector field by approximating its spatial derivatives on the Cartesian grid using centered finite difference (see details in Supplementary Materials Section 2.1). The resulting multi-channel representation consists of three decomposed components from each of the three RGB channels, producing nine feature channels that are concatenated and used as input to the downstream CCNN.

\subsection{Combinatorial Complex Neural Network Overview }
The proposed CCNN architecture operates on a combinatorial cell complex $\mathcal{X}$, enabling structured message passing among multiple cell dimensions. A combinatorial complex can be viewed as a collection of cells supported on subsets of image grid entities. A rank function assigns a non-negative integer to each cell, indicating its geometric dimension and hierarchically organizes the cells. In our approach, each decomposed image defined on a Cartesian grid induces a two-dimensional combinatorial complex consisting of $0$-cells and $2$-cells. 

The $0$-cells serves as spatial sampling points of the image, while the $2$-cells correspond to square regions formed by adjacent $0$-cells. An $n$-cochain is viewed as a feature vector defined on the $n$-cell of the combinatorial complex \cite{hajij2022topological}. We denote the associated cochain spaces by $C^0(\mathcal{X})$ and $C^2(\mathcal{X})$. Within each cell dimension, neighborhood operators establish adjacency relations that create discrete cochain maps for higher-order message passing. Higher-order message passing may be considered as a computing framework in which information is passed between $0$-cells and $2$-cells using neighborhood operators.

\subsubsection{Combinatorial Complex representation of Images}
Consider $\hat{I}\in \mathbb{R}^{h \times w \times \Tilde{c} }$ as the concatenated multi-channel image obtained after the decomposition (See Supporting Information S2) where $\Tilde{c}=9$. We construct $0$-cells and $2$-cells from $\hat{I}$. The $0$-cells are created through a patch embedding process, which is executed as a convolution with kernel size of $p$ and stride of $p$. This results in a regular grid with dimensions $h_{p} \!\times\! w_{p}$ where $w_{p}=\frac{w}{p}$ and $h_{p}=\frac{h}{p}$. Each patch location $(i,j)$ corresponds to a $0$-cell that is associated with a feature vector $\mathcal{C}^{0}(\mathcal{X})$. We define adjacency among $0$-cells using an $8$-adjancency scheme. For every position $(i,j)$, we collect all horizontal, vertical and diagonal neighbors lying within the grid domain. This is achieved through an offset set 
\begin{equation}
    \{(-1,-1),(-1,0), (-1,1), (0,-1), (0,1), (1,-1),(1,0), (1,1)\},
\end{equation}
which defines the set of edges in the $0$-cell graph.

The $2$-cells are constructed by applying a $2p \!\times\! 2p$ convolution with stride of $p$, where each output is a square formed by four neighboring $0$-cells. This operation can be seen as a parametrized pushforward from $\mathcal{C}^{0}(\mathcal{X})$ to $\mathcal{C}^{2}(\mathcal{X})$ with $0$-cell information aggregated to define square regions. The connectivity among $2$-cells follows a $4$-adjacency structure defined by the offsets
\begin{equation}
    \{(-1,0),(1,0),(0,-1),(0,1)\},
\end{equation}
defining the adjacency graph on the $2$-cells.

\begin{figure}[htbp]
    \centering
    \includegraphics[width=1\linewidth]{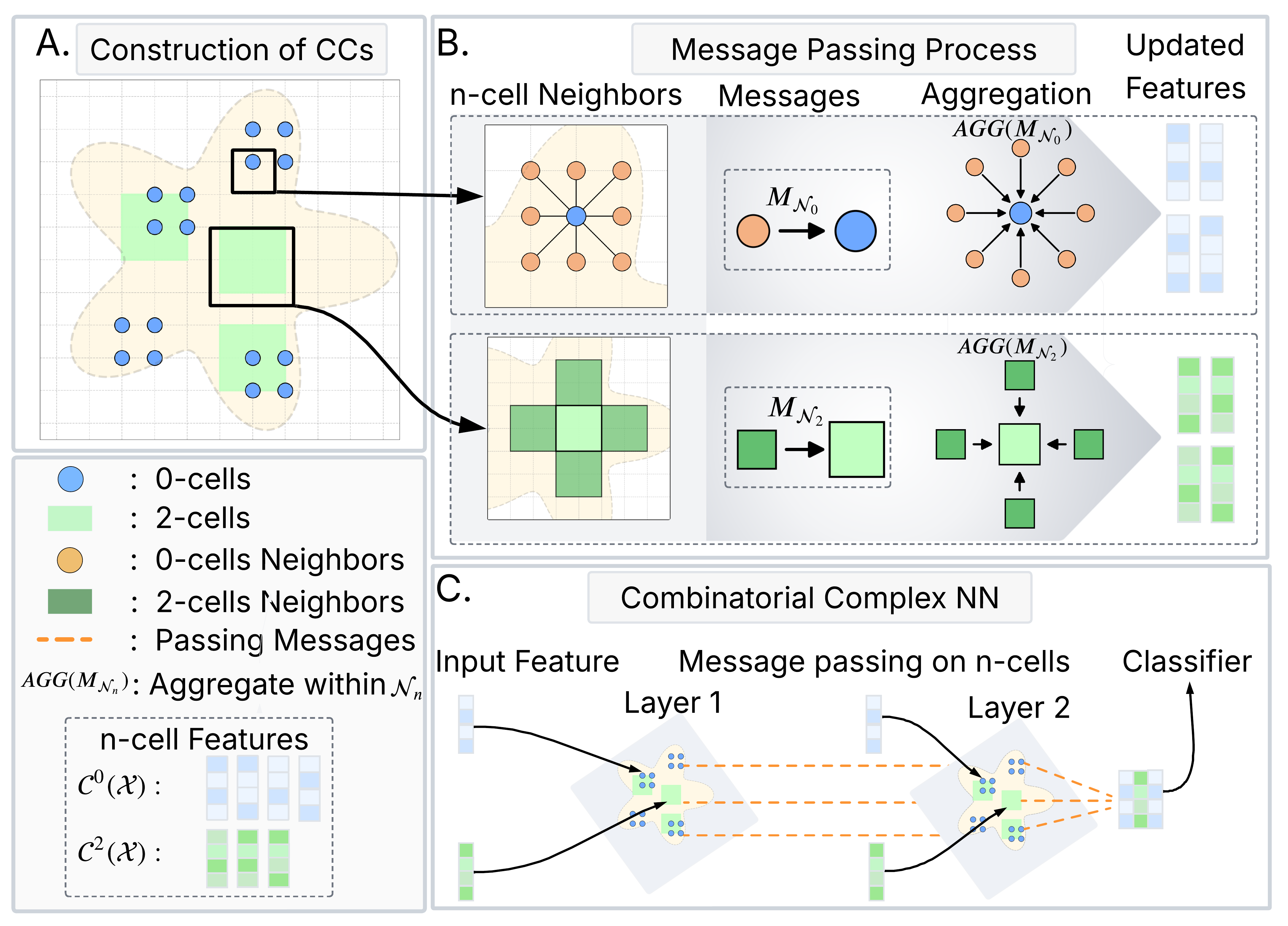}
    \caption{Illustration of Combinatorial Complex Neural Network. (A) Combinatorial Complex is constructed on the star domain using $0$-cells (nodes) and $2$-cells (faces). (B) Neighborhood message passing is performed on each $n$-cell and message aggregation performed within the adjacent neighbors. (C) The resulting features are propagated through successive CCNN layers, fused and passed to a classifier for the final task.}
    \label{fig:cc_flowchart}
\end{figure}

Fig.~\ref{fig:cc_flowchart} illustrates a setup for the CCNN. A combinatorial complex is constructed on the star shaped domain as shown in Fig.~\ref{fig:cc_flowchart}A consisting of $0$-cells and $2$-cells. In Fig.~\ref{fig:cc_flowchart}B, adjacency neighborhoods associated to the target $n$-cell are defined by $\mathcal{N}_0$ and $\mathcal{N}_2$ denoting $0$-cell and $2$-cell neighbors respectively. Messages $M_{\mathcal{N}_{0}}$ and $M_{\mathcal{N}_{2}}$ associated with each neighbor $\mathcal{N}_{0}$ and $\mathcal{N}_{2}$ respectively are collected separately and later the messages across all neighbors aggregated for the target $n$-cell. The feature on the target $n$-cell is then updated. In particular, for the $0$-cell branch in Fig.~\ref{fig:cc_flowchart}B (top part), $8$ adjacent neighbors for a target $0$-cell are identified. The corresponding messages $M_{\mathcal{N}_{0}}$ are first collected from adjacent neighboring $0$-cells and then aggregated to update the target $0$-cell feature. Fig.~\ref{fig:cc_flowchart}C presents the overall CCNN pipeline, where the initial $n$-cell features are defined on the constructed combinatorial complex and propagated through successive message passing layers. Within each layer, neighborhood aggregation is performed on the cell neighborhoods yielding updated cell representations. By stacking multiple CCNN layers, the network progressively captures richer structural information from the complex. The final updated cell features are then combined and passed to a classifier.

\subsubsection{Implementation of an n-cell Branch }
 For cell dimension $n\in \{0,2\}$, the corresponding feature map is obtained via convolutional embedding. For $n=0$, we apply a convolution with kernel and stride equal to $p$. For $n=2$, we use kernel size $2p$ and stride $p$, which aggregates neighboring $0$-cells to produce $2$-cell features. In both cases, the convolution simultaneously performs patch embedding and linear projection
\begin{equation}     
Z_{n\text{-cell}}= \text{Conv}_{p}(\hat{I}), \quad Z_{n\text{-cell}}\in \mathbb{R}^{b \times d \times h_{n} \times w_{n}}, 
\end{equation} 
where $d$ is the embedding dimension, $b$ is the batch size, $w_{n}$ and  $h_{n}$ denote the resolution of the $n$-cell. The feature map obtained is reshaped into a sequence of $N_{n}$ tokens
\begin{equation}
    X_{n\text{-cell}}\in \mathbb{R}^{b \times N_{n} \times d},
\end{equation}
where $N_{n}=h_{n}w_{n}$. The $n$-cell combinatorial graph, denoted as $G_{n}$ is constructed from the corresponding neighborhood function. Graph attention layers with residual connections are used to implement intra-neighborhood message passing within each cell rank. We stack two attention and feedforward blocks in each combinatorial phase. For each update, we have 
\begin{align}
   & X^{1}_{n\text{-cell}} = X_{n\text{-cell}} + \text{Drop}(\text{GAT}(\text{BatchNorm}(X_{n\text{-cell}}), G_{n})), \\ 
   & X^{2}_{n\text{-cell}} = X_{n\text{-cell}}^{1} + \text{Drop}(\text{FFN}(\text{BatchNorm}(X^{1}_{n\text{-cell}}))),
\end{align} 
where GAT is the graph attention operator and FFN is a feed-forward network with GELU activation. The new feature is reshaped back into its spatial form 
\begin{equation}
    \hat{Z}_{n\text{-cell}}\in \mathbb{R}^{b\times d \times h_{n}\times w_{n}},
\end{equation}
 and mapped to the image grid by a transposed convolution. Finally, max pooling is applied to gradually reduce the spatial resolution. The spatial downsampling can be interpreted as a combinatorial complex pooling. Local features are aggregated into a coarser representation supported on a higher-level cell.
 
The feature dimensions from both branches are aligned through adaptive pooling. In the final stage, the updated features are fused through a learnable gate mechanism and later concatenated. A sigmoid activation function is used to constrain gating coefficients within the range of $(0,1)$. The fused features are obtained by the element-wise combination of the two branches. A global average pooling is performed to obtain a compact feature vector that is passed to a linear classifier for the image classification tasks.

\subsection{Evaluation Metrics}\label{subsec:metrics}
We evaluate model performance using classification accuracy (ACC) and the Area Under the Receiver Operating Characteristic Curve (AUC)\cite{bradley1997use,rosset2004model}. ACC measures the proportion of correctly classified samples based on discrete predictions obtained via thresholding or the argmax operation. AUC assesses the discriminative ability of the model by evaluating continuous prediction scores in a threshold independent manner \cite{ccorbaciouglu2023receiver,yang2023medmnist}. All experiments were conducted using five different random seeds and the average AUC and ACC obtained from these runs are presented as the final result for comparison with the benchmark models. 

For datasets with binary classification, multi-class and ordinal regression tasks we used the cross-entropy loss
function. CNMTDL is implemented in PyTorch \cite{paszke2019pytorch} and trained on NVIDIA H200 GPU with CUDA 12.6
acceleration. More details on the model parameters, loss function and benchmark models are provided in the Supplementary Materials (Supplementary Section 4).

\section*{Supplementary Materials}\label{sec: SI}
 Additional materials are provided in the Supporting Materials, including related work, methodological details, dataset description, model parameterization, dataset analysis, and performance analysis.  
 
\section*{Conflicts of Interest}
The authors declare no competing financial interests

\section*{Data Availability}
The dataset used is available online via Zenodo at \url{https://zenodo.org/records/10519652}.

\section*{Code Availability}
The codes used can be found in Github at \url{https://github.com/wachiraa26/CNMTDL}.

\section*{Acknowledgments}
 This work was supported in part by NIH grants R01AI164266 and R35GM148196, MSU Research Foundation, The University of Georgia, and Georgia Research Alliance.   
%
\bibliographystyle{unsrt}
\bibliography{references}

\end{document}